# Data-Driven Design: Exploring new Structural Forms using Machine Learning and Graphic Statics

Lukas FUHRIMANN [a], Vahid MOOSAVI [b], Patrick Ole OHLBROCK [a], Pierluigi D'ACUNTO [a]

[a] ETH Zurich, Institute of Technology in Architecture, Chair of Structural Design
Stefano-Franscini-Platz 5, 8093 Zürich (CH)
lukasf@ethz.ch

[b] ETH Zurich, ITA, Chair for Computer Aided Architectural Design

**Abstract**

The aim of this research is to introduce a novel structural design process that allows architects and engineers to extend their typical design space horizon and thereby promoting the idea of creativity in structural design. The theoretical base of this work builds on the combination of structural form-finding and state-of-the-art machine learning algorithms. In the first step of the process, Combinatorial Equilibrium Modelling (CEM) is used to generate a large variety of spatial networks in equilibrium for given input parameters. In the second step, these networks are clustered and represented in a form-map through the implementation of a Self Organizing Map (SOM) algorithm. In the third step, the solution space is interpreted with the help of a Uniform Manifold Approximation and Projection algorithm (UMAP). This allows gaining important insights in the structure of the solution space. A specific case study is used to illustrate how the infinite equilibrium states of a given topology can be defined and represented by clusters. Furthermore, three classes, related to the non-linear interaction between the input parameters and the form space, are verified and a statement about the entire manifold of the solution space of the case study is made. To conclude, this work presents an innovative approach on how the manifold of a solution space can be grasped with a minimum amount of data and how to operate within the manifold in order to increase the diversity of solutions.

## 1. Introduction

Focusing on equilibrium-based design can help architects and engineers to explore original structural forms in the conceptual design phase. Novel computationally-driven methods such as Combinatorial Equilibrium Modelling (CEM) (Ohlbrock et al. [1]) based on graphic statics, can potentially generate an enormous number of spatial networks in equilibrium in a relatively short amount of time. It is clear, however, that a human designer has no chance to control manually the entire solution space generated by such an algorithm. At most, the designer can only explore a small subset of it, which is typically obtained through a sequential navigation within the solution space. This subset can be regarded as an individual design horizon, which is not only limited in size, but also in quality because it is heavily influenced by the individual experience and design references in the mind of the designer.

The present work tries to overcome these quantitative and qualitative limitations, allowing architects and engineers to go beyond their individual approachable design horizons and thereby supporting the idea of diversity-driven design (Brown and Mueller [2]). This should support the designer in an early phase, either to explore various solutions for an abstract study or to find design variations within a defined design brief. The theoretical base of this work builds on the combination of a form-finding methodology and state-of-the-art machine learning algorithms. Previous works such as Brown and Mueller [2], Mueller and Ochsendorf [3] or Shea and Cagan [4] have shown the great potential of data-driven structural design.

Thanks to the computational inexpensiveness in the generation of spatial networks in equilibrium (in this work also referred as *equilibrium forms*) and due to the possibility to combine compression and tension elements in the same structural form, CEM is used in the present work as an efficient form-finding tool for the construction of the design space. In order to represent the infinite solution space of equilibrated forms that can be produced for a given set of input parameters, a large but finite amount of data can be used. Although being a subset, this data can provide a diverse representation of the solution space. Machine learning algorithms such as Self Organizing Map (SOM) (Kohonen [5]) and Uniform Manifold Approximation and Projection (UMAP) (McInnes and Healy [6]) are then implemented to analyze this design space and control the interaction between input parameters and resulting equilibrium network.

## 2. Combinatorial Equilibrium Modelling

CEM is a method for the design of spatial networks in equilibrium that is based on graphic statics. This allows for the generation of a design space with infinite different equilibrium states that can be conveniently represented using a *form diagram* **F** and a *force diagram* **F\*** (Ohlbrock et al. [1]). Other than the conventional form and force diagrams, CEM introduces a generative *topological diagram* **T**, which enables the control of the load-bearing behavior of a network in equilibrium by varying the connectivity of the network, its combinatorial state (tension or compression), the length of its elements as well as the magnitudes of its inner forces. Within the CEM framework, the members along the shortest paths from each vertex to the closest support vertex are defined as *trail members* while the others *deviation members*. Start vertices $s$ are those vertices of a trail with the maximum topological distance $w$ from their corresponding support vertices. The topological diagram is split into layers $k$ according to the topological distance $w$ of the individual vertices to their corresponding support vertices. (Ohlbrock and Schwartz [7]).

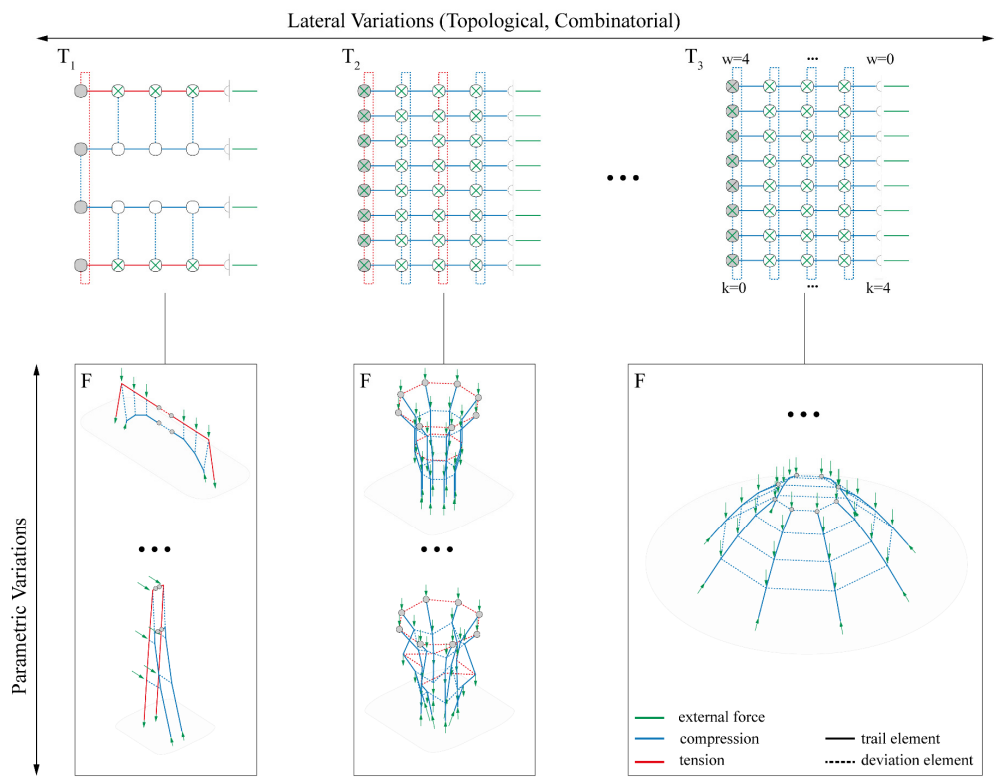

Figure 1: CEM: Schematic representation of lateral and vertical design variations

Figure 1 shows three different variations of an initial topological diagram (*lateral variation*), each of them spanning a distinguished solution space. By modifying the input parameters of each individual topological diagram (*vertical variation*), an infinite range of equilibrium states within these spaces can be created. For example, the topological diagram $T_3$ (right side of Figure 1) has 64 degrees of freedom in total: 32 parameters defining the inner force magnitudes of the deviation members and 32 parameters defining the lengths of the trail members. For a network that can be represented by a simple topology (up to 20 elements), the nonlinear behavior of the overall equilibrium state, which is due to the sequential interaction between the equilibrium of the different layers of the structure, can be easily controlled by a human designer. On the contrary, when increasing the complexity and the amount of elements of the network, the nonlinear behavior of the equilibrium state - and therefore the relation between parameters and resulting equilibrium shapes within the design space - is increasingly difficult to predict. In the following, an approach is suggested for the vertical exploration of these cases of complex structures.

### 2.1. Topological Set-Up

In order to illustrate the proposed methodology, an exemplary topological diagram is taken into account. The investigation of this topological diagram allows conducting an abstract study of forms where no defined design brief is specified. Nevertheless, the same general methodology can be applied to design studies bounded to a defined design brief.

The herein investigated topological diagram is based on the diagram $T_3$ of Figure 1, with an expanded number of 400 trail and 400 deviation members. For this topology, an equilibrium state can be obtained once a deviation force magnitude $D$ is assigned to each deviation member and a constraint plane $P$ is defined for each node (Ohlbrock et al. [1]). These planes fix the position of the nodes of the structure in space and thus control the lengths of the trail members. External loads, which represent the self-weight of the structure, are also applied at every node of the network. Within this study, the positions of the start vertices $s$ are fixed and equally distributed on a circle with a variable radius $R$. This topological setup thus results in a total amount of 400 parameters describing the deviation forces $D$, 420 parameters for the constraint planes $P$ and one parameter describing the radius $R$.

### 2.2. Parametric Set-Up

In order to reduce the number of parameters and therefore obtain a more systematic control over the generated spatial networks in equilibrium, relationships between input parameters are established by means of mathematical functions (Figure 2). Specifically, sinusoidal functions are used to describe the relationship between parameters within the same layer $k$ of the structure, but also to define the correlation between amplitudes over the various layers. Sinusoidal functions are chosen due to their continuity and smoothness. The parameters $A_A$, $A_B$, $A_C$, $A_H$, $A_D$ and $E$ describe the shift of the function, the parameters $B_A$, $B_B$, $B_C$, $B_D$, $B_H$, $F_E$ and $F$ describe the amplitude of the function and the parameters $C_A$, $C_B$, $C_C$, $C_D$, $C_H$, $G_E$ and $G$ describe the frequency.

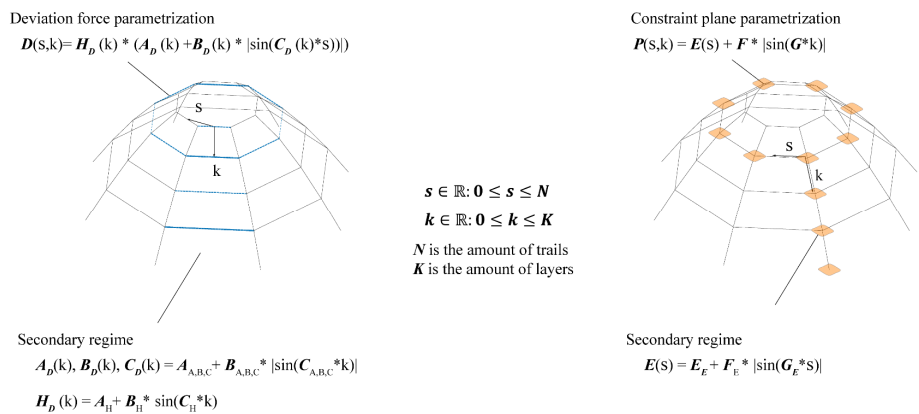

Figure 2: Overview on the parametric set-up

Thanks to the use of sinusoidal functions, the 821 input parameters can be reduced to 18 only. Additionally, the external forces applied to the nodes of the first layer *L* (where *L=D*(s,0)) are kept constant and can be represented with an additional single parameter. Hence, a total amount of 19 parameters is used to describe the entire solution space. Restricting the values to integer numbers, each ranging between 0-100, a total of **$100^{19}$** different combinations can be found. An automated generation of forms is implemented, where these 19 input parameters are randomly chosen.

### 2.3. Form Generation and Evaluation

Forms generated with CEM are always in static equilibrium but obviously not all of them represent meaningful structural forms. Within this study, a simple geometric criterion is formulated in order to determine whether the generated equilibrium form is accepted or not. In particular, every single sequence of each structure is checked for projected intersections. The structure is discarded, if an intersection between the elements of one or more layers is detected. In total, 252'000 different equilibrium networks have been generated and analyzed within the scope of this project. The application of the geometric criterion decreased the number to 35'800 accepted equilibrated forms for which the SOM and UMAP algorithms have been subsequently trained.

## 3. Self Organizing Map and Uniform Manifold Approximation and Projection

Self Organizing Map (SOM) is a powerful nonlinear manifold learning method introduced by Kohonen [5]. From a mathematical point of view, SOM transforms (*maps*) the data from the originally high-dimensional space to a low-dimensional space (usually a space of two dimensions), while the topology of the original high dimensional space is preserved. Topology preservation means that if two data points are similar in the high-dimensional space, they are necessarily close in the new low-dimensional space. This low-dimensional space is normally represented by a planar grid with a fixed number of points, which is defined as map. Each node of this map has its own coordinates ($x_{i,1}$, $x_{i,2}$) and an associated high-dimensional vector $W_i = \{w_{i1}, \dots, w_{in}\}$, where the original observed data is represented by *n* dimensional vectors. In comparison to other data representation methods, SOM has the advantage of delivering two-dimensional maps visualizing smoothly changing patterns of data from the original high-dimensional space. This makes SOM a very intuitive method for dimensionality reduction, data clustering and visualization. In addition, SOM can also be used for nonlinear approximation functions (Barreto and Souza [8]).

Uniform Manifold Approximation and Projection (UMAP) is a dimension reduction technique similar to well-known methods such as t-SNE (Van der Maaten and Hinton, [9]) with additional benefit of scalability to large data sets. The algorithm models each high-dimensional object by a two-dimensional point in such a way that similar objects in the high dimensional space are represented by nearby points in the two-dimensional space and dissimilar objects are modeled by distant points. In comparison to SOM, UMAP is not restricted to a fixed number of points in a planar grid and allows representing all input data in an open space, which is beneficial for the representation of clusters.

The dimension *n* of the input vectors of a SOM and UMAP is determined by the clustering objective of the designer. A recent example in the context of structural design was presented by Harding [10] in which 3 objectives were chosen (twist, height and taper) to represent a solution space of towers with a SOM. In the current study, the networks generated with the CEM can be described by their nodal coordinates (n=1260). The nodal coordinates are also the input for the training of the SOM algorithm. The nodes of each input are normalized in each direction, in order to allow comparing the variation of a form in relation to all the others. Hence, the objective is similarity of form diagrams in Euclidean space (i.e. Euclidean distances between corresponding nodes) expressed through 1260 coordinates.

## 4. Solution Spaces, Self Organizing Maps and Uniform Manifold Approximation and Projection

Each accepted equilibrium state generated by the CEM (Section 2.3) is abstracted into a single input vector containing the nodal coordinates of the corresponding form diagram. The whole vector therefore contains 1260 resulting nodal coordinates.

Within this study, SOM is used to visualize and cluster the abstracted data. The output is a structured form-map, in which the designer can interactively move in the clustered design space of spatial equilibrium networks. In order to explain how this design space can be extended and how different regions in this map can be described in more depth, the correlation between the input parameter and the emerging clusters of forms is identified with the UMAP algorithm. The experiments were implemented in Python environment, using SOMPY (Moosavi [11]) and UMAP (McInnes [6]).

### 4.1. SOM of Equilibrium Structures

In a first step, the application of the SOM algorithm results in a fixed grid of nodes and weights, with exemplary instances of the forms in equilibrium taken from the input dataset (in this case a 80x80 grid 6400 nodes). It is important to note that the nodes of the SOM are trained with the input data but do not directly represent those. In a second step, the form-map is created (Figure 3 center), in which every node directly represents at least one given input. This is done by assigning each of the inputs to a node within the SOM grid, which has the best matching SOM representation. In case no node of the SOM grid is found to match a given input, a gap is generated in the form-map. This results in a clustered representation of all inputs, where similar forms are found within one node (*form families*) and clusters of forms are delineated by the gaps in the map. Interactively moving within this map of spatially equilibrated forms allows the exploration of potential structures that could hardly be designed manually. Four exemplarily selected forms, indicating the diversity of the solution space, are shown in Figure 3: a twisted tower (bottom left), a radially symmetrical tower with a gradually varying floor section (top left), a shell in tension and compression (top right) and a dome in pure compression (bottom right). It is important to keep in mind that these forms represent four different realizations of the same given topology.

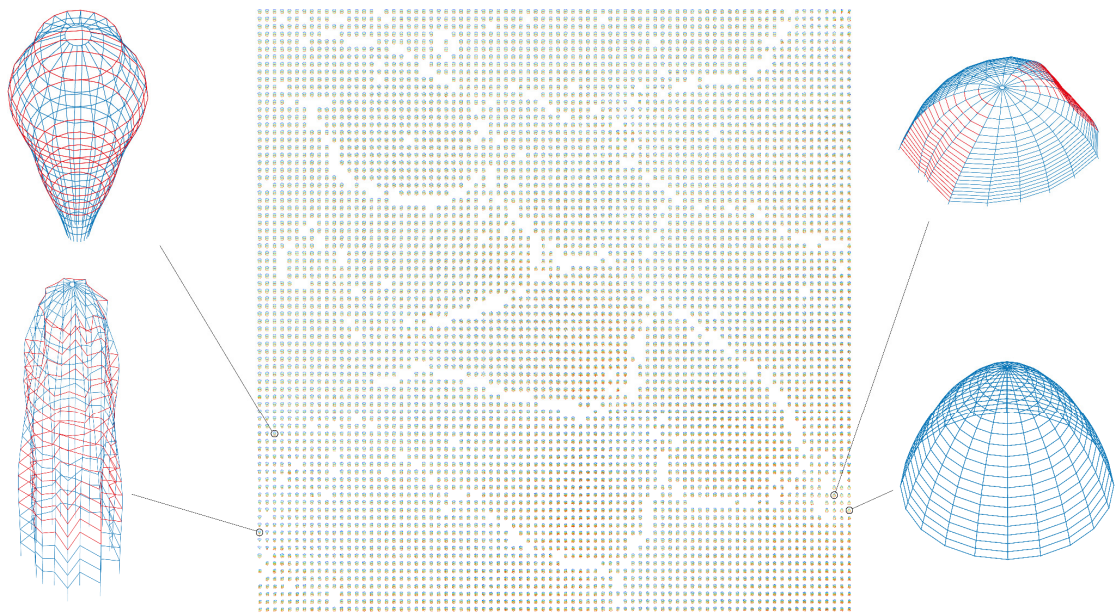

Figure 3: Representation of the form-map (center) and four exemplary forms selected from the map

To understand the map in a more detailed resolution, three examples of form families are selected and shown in Figure 4, which are respectively assigned to three different nodes in the map.

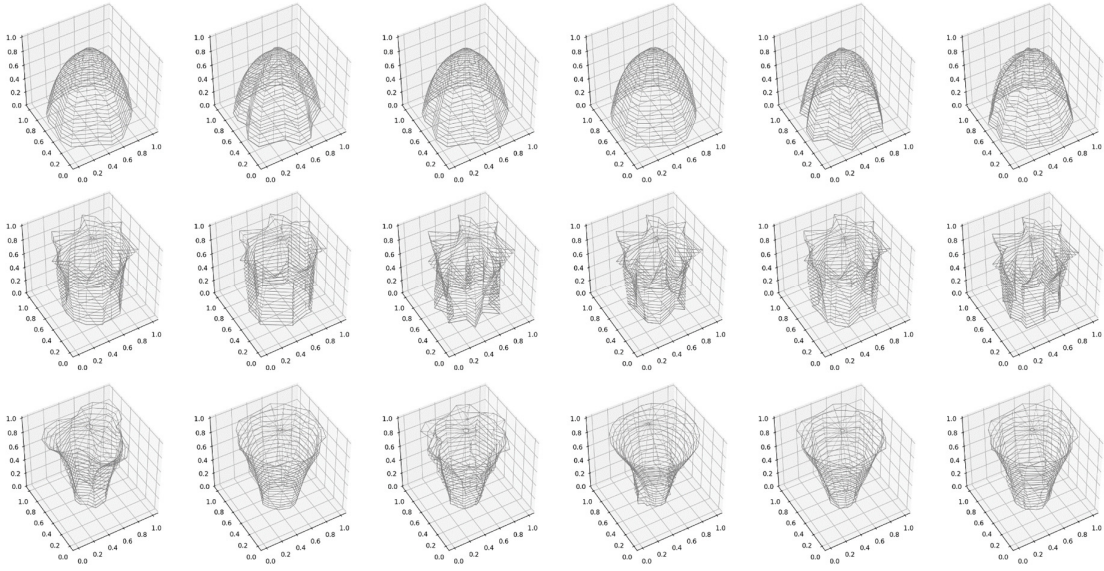

Figure 4: Representation of three different form families from the form-map

### 3.4. UMAP Representation of Cluster and CEM Parameters

In order to better distinguish the different clusters, the trained data of the UMAP algorithm is analyzed in a second step. Each point of the scatter plot in Figure 5 (left) corresponds to a single input form, which is modelled as a two-dimensional point by the algorithm. The point clouds in Figure 5 indicate similar forms in the Euclidean space and the distance between the clouds shows the metric resemblance between the clusters. Nine different clusters can clearly be distinguished.

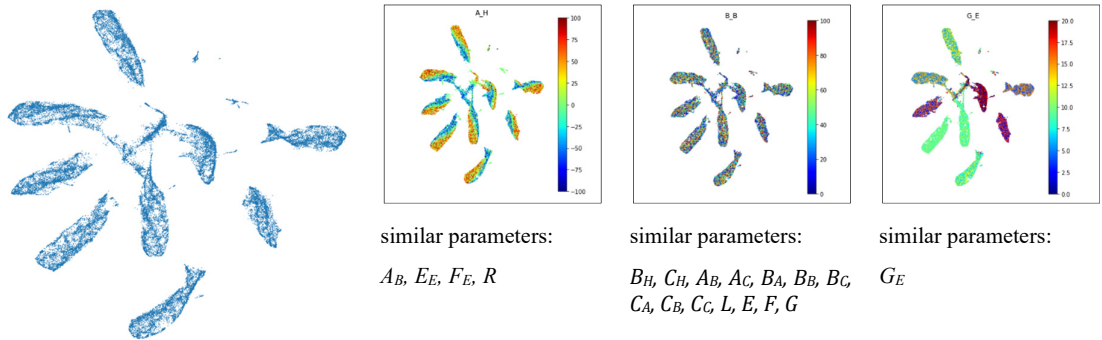

similar parameters:

$A_B$, $E_E$, $F_E$, $R$

similar parameters:

$B_H$, $C_H$, $A_B$, $A_C$, $B_A$, $B_B$, $B_C$, $C_A$, $C_B$, $C_C$, $L$, $E$, $F$, $G$

similar parameters:

$G_E$

Figure 5: Clustered space of UMAP (left) and UMAP parameter representation of three different classes

To understand the correlation between the CEM input parameters and the generated equilibrium forms, a visualization of the different parameters for the trained UMAP is carried out in a third step. Three qualitatively different classes of parameter correlation can be detected, as shown on the right side of Figure 5. Parameters that direct the resulting equilibrium form within a cluster into a specific region (left), parameters that are arbitrary distributed within each cluster (middle) and parameters that clearly describe to which cluster an equilibrium form belongs (right).

Two sensitivity studies are finally presented and discussed, both trying to prove the distinguished classes of the parameter correlations. Firstly, a new set of data is produced with a constant parameter $A_B = 5$ to empirically validate that this parameter has no influence in the overall variation of equilibrium forms that can be found. Secondly, a new set of data is produced with a reduced numeric range of randomly distributed numbers of parameter $G_E$ between 8 and 12. This should validate that parameter $G_E$ controls the amount of clusters generated. Both data sets are trained with UMAP and the corresponding scatterplots are shown in Figure 5.

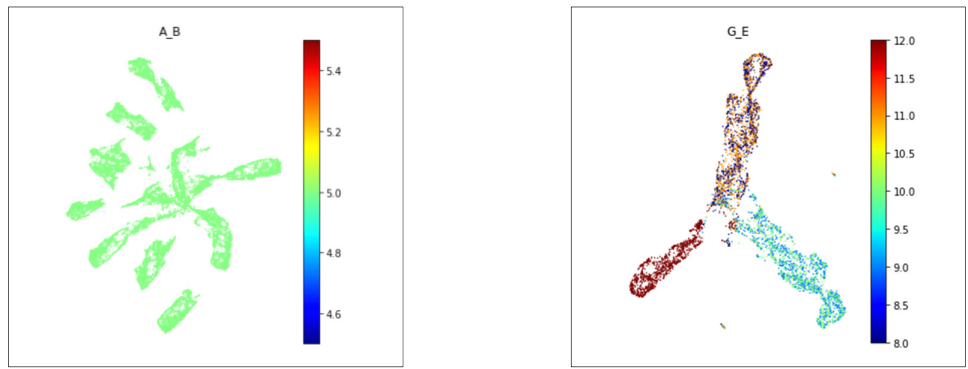

Figure 6: UMAP parameter representation for constant parameter $A_B = 5$ (left) and reduced parameter range (8-12) for $G_E$ (right).

The left representation of Figure 6 shows the clustered space for a constant parameter $A_B = 5$. The same amount of clusters exists in Figure 5. This proofs the first statement, which says that no correlation between the equilibrium forms and this parameter exists. This statement can be extended to all parameters that describe the similar behavior in Figure 5, which results in five defining parameters for the represented design space. Figure 6 (right) shows the three different clusters forming in the new parameter space where $G_E$ is varying in a reduced numeric range. Also in Figure 5 only three clear clusters can be detected for the reduced numeric range (in turquoise). This proofs the second statement, which says that parameter $G_E$ defines the amount of clusters in the solution space. Increasing the parameter to values higher than 20 would not change the clustered space in Figure 5 because the parameter represents the frequency of a sinusoid and its discretization resolution is restricted by the amount of trails (20 for the defined topology). This indicates that the entire diversity of clustered space is represented for the current case study.

## 4. Conclusion

The present research showed the potential of combining a simple and fast form-finding technique with machine learning. The paper presented strategies to generate, represent and evaluate a large solution space of novel equilibrium networks by combining CEM with SOM and UMAP. After the methodology is described, a case study is used to show the potential of this novel design approach.

A form-map, in which the funicular geometries are clustered according to their variation in the Euclidean space, allows the designer to explore an infinite, yet ordered formal variation of spatial equilibrium networks (Figure 7, "city of equilibrium forms"). Combining the aforementioned approaches, this eventually enables structural designers in an early design phase to reach a more creative, diversity-driven and structurally informed design. Depending on the concerns of the designer, the clustering objectives of the map can be altered or enlarged (e.g. objective could be the total height, the width, the weight or the inner forces of the equilibrium network). An interaction with the form-map additionally enables the designer to investigate/study interesting areas in the map and create a more detailed representation of these.

To further understand and interpret the design space, the UMAP algorithm is used. The application to the case study results in nine clusters of equilibrated forms. Based on the developed connection between the CEM parameter representation and the solution space, three different correlation classes can be distinguished. One parameter defines the amount of clusters, four parameters define the region within a cluster and fourteen parameters do not have an influence on the diversity of the solution space. Altogether, this enables a clear understanding of the entire, infinite solution space for the given topological and parametric setup in this project. As an example, the creation of a data set in different numeric ranges can further increase the size of a cluster; however, the entire amount of equilibrium network clusters will not change. To conclude, this work presents an innovative approach on how the manifold of a solution space can be grasped with a minimum amount of data required. Knowing the manifold of a solution space, allows to change the latter and, by this, to increase the diversity of solutions. This concept will be further investigated in future research.

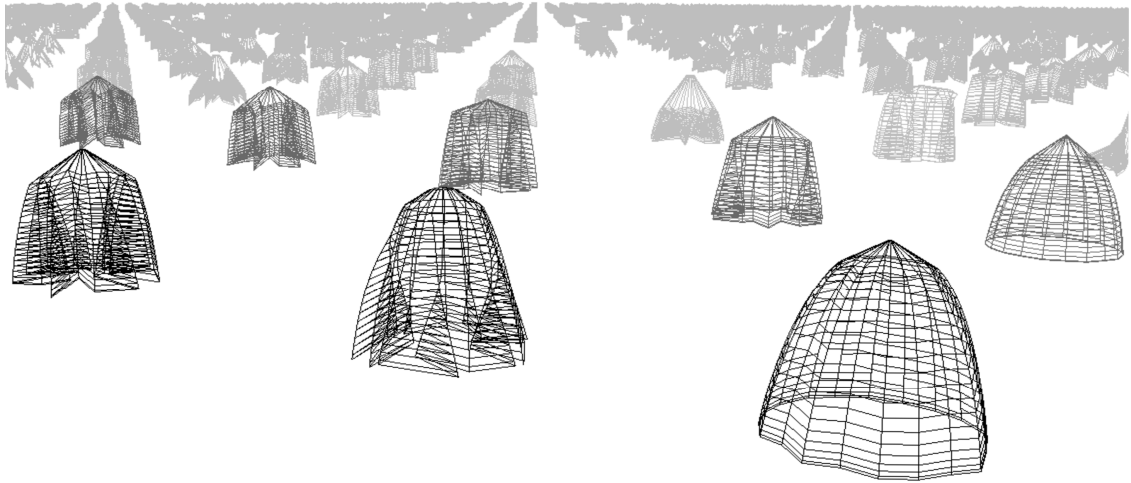

Figure 7: Three-dimensional representation of the form-map as a "city of equilibrium forms"